\documentclass[10pt,twocolumn,letterpaper]{article}

\usepackage{cvpr}              
\usepackage[accsupp]{axessibility}

\usepackage{tikz}
\usepackage{multirow}
\usetikzlibrary{shapes, arrows, positioning}

\usepackage{graphicx}
\usepackage{tikz}
\usepackage{subcaption}
\usepackage{amsmath}
\usepackage{tabularx}  
\usepackage[utf8]{inputenc}

\definecolor{cvprblue}{rgb}{0.21,0.49,0.74}
\usepackage[pagebackref,breaklinks,colorlinks,allcolors=cvprblue]{hyperref}

\def\paperID{36} 
\def\confName{CVPR}
\def\confYear{2025}

\title{PhysNav-DG: A Novel Adaptive Framework for Robust VLM-Sensor Fusion in Navigation Applications}

\author{
Trisanth Srinivasan \quad Santosh Patapati \\
Cyrion Labs \\
{\tt\small \{trisanth, santosh\}@cyrionlabs.org}
}

\begin{document}
\maketitle
\begin{abstract}
Robust navigation in diverse environments and domains requires both accurate state estimation and transparent decision making. We present PhysNav-DG, a novel framework that integrates classical sensor fusion with the semantic power of vision-language models. Our dual-branch architecture predicts navigation actions from multi-sensor inputs while simultaneously generating detailed chain-of-thought explanations. A modified Adaptive Kalman Filter dynamically adjusts its noise parameters based on environmental context. It leverages several streams of raw sensor data along with semantic insights from models such as LLaMA 3.2 11B and BLIP-2. To evaluate our approach, we introduce the MD-NEX Benchmark, a novel multi-domain dataset that unifies indoor navigation, autonomous driving, and social navigation tasks with ground-truth actions and human-validated explanations. Extensive experiments and ablations show that PhysNav-DG improves navigation success rates by over 20\% and achieves high efficiency, with explanations that are both highly grounded (score 0.87) and clear (readability 4.5/5). This work connects high-level semantic reasoning and geometric planning for safer and more trustworthy autonomous systems. 
\end{abstract}    
\section{Introduction}
\label{sec:intro}

Autonomous navigation in real-world environments poses a fundamental challenge: an agent should generalize across domains (e.g., indoor corridors, outdoor roads, crowded pedestrian scenes) while making decisions that are understandable and trustworthy to humans. Classic robotics approaches achieve reliability via sensor fusion and filtering (e.g., Kalman filters \cite{kalman1960} to fuse lidar, radar, and odometry for accurate state estimation), but they lack semantic understanding and explainability. On the other hand, modern vision-language models (VLMs) such as CLIP \cite{radford2021clip} and GPT-4V \cite{openai2023gpt4} encapsulate rich visual-semantic knowledge that could improve an agent’s reasoning and adaptability across domains. However, integrating these large pre-trained models into an embodied navigation system, and doing so in a way that yields real-time explanations for the agent’s actions, remains an open problem.

\medskip

\noindent Explainability is critical for safety and trust in autonomous systems \cite{kuznietsov2024xai, atakishiyev2025safetyxai}. A human co-driver or robot operator should be able to understand vehicle actions and robot planning. Traditional navigation pipelines and end-to-end policies rarely offer human-readable justifications. Recent advances in large language models suggest that forcing an AI agent to ``think out loud'' in natural language can both improve decision quality and provide an intrinsic explanation for each action. In the context of robotics, Zawalski \emph{et al.} \cite{zawalski2024ecot} showed that an embodied agent with explicit chain-of-thought reasoning improved success rates by up to 28\% on challenging tasks. These developments motivate our approach: we combine classical low-level sensor fusion with high-level vision-language reasoning to create a navigation agent that both performs well across domains and explains its decisions in real time.

\medskip

In this paper, we introduce a dual-branch VLM framework for explainable, domain-adaptive navigation. Our system consists of two synchronized branches: one branch processes multi-modal sensor inputs to predict navigation actions, and a parallel branch generates a natural language explanation for each decision. The two branches share information and are trained with a consistency loss to ensure the explanations align with the actions. We further enhance reliability through an adaptive sensor fusion mechanism: the classical state estimator and controller (e.g., a Kalman-filtered motion planner) is fused with the VLM policy outputs. This allows the system to fall back on proven geometric planning when needed, while leveraging the VLM’s semantic insight when it is most beneficial. A feedback loop uses explanation fidelity checks to adjust this fusion in real time, ensuring that when the VLM says it is reacting to a certain obstacle or instruction, the fusion logic verifies and trusts the VLM appropriately.

\medskip

To rigorously evaluate our approach, we developed the Multi-Domain Navigation and Explanation Benchmark (MD-NEX). Existing benchmarks tend to focus on single domains or either performance or explanations alone. For example, Room-to-Room indoor navigation \cite{ku2020rxr} tests language instruction following, and the BDD-X driving dataset \cite{kim2018bddx} provides human explanations for driving, but no unified benchmark covers multiple environment types and explainability. Our benchmark spans indoor navigation 
(in simulated homes/offices using Matterport3D \cite{chang2017matterport3d}), 
outdoor driving 
(with urban driving scenarios from BDD-X, augmented by sensor data from nuScenes \cite{caesar2020nuscenes} and Waymo \cite{sun2020waymo}), 
and social navigation in pedestrian-rich environments
(building on the recent SNEI dataset \cite{liu2025snei} for socially compliant navigation with explanations). 
For each domain, it includes diverse scenarios with ground-truth goal completion criteria and ground-truth explanation annotations, enabling quantitative evaluation of both navigation success and explanation quality.

\medskip

\noindent In summary, our contributions are fourfold:
\begin{enumerate}[label=(\alph*)]
    \item \textbf{Dual-Branch VLM Framework:} We propose a novel architecture that integrates a vision-language model into a navigation agent via two coordinated branches, one for low-level action outputs and one for high-level natural language reasoning, providing real-time chain-of-thought explanations without negatively impacting performance.
    \item \textbf{Adaptive Sensor Fusion:} We design an adaptive fusion mechanism that combines classical sensor-fusion-based planning with VLM-predicted actions. A dynamic weighting scheme, informed by confidence estimates and explanation fidelity via an erasure check, selects the optimal blend of VLM guidance and reliable geometric control at each time step.
    \item \textbf{Multi-Domain Navigation-Explanation Benchmark:} We construct a comprehensive benchmark covering indoor, outdoor, and social navigation tasks, each paired with explanation annotations. The dataset is compiled from and extends existing sources (Matterport3D/R2R, BDD-X with added sensor streams, SNEI), supplemented by automatically generated scenarios, and includes standardized evaluation metrics for both navigation (SR, SPL, collision rate) and explanation (faithfulness, readability, consistency).
    \item \textbf{Extensive Experiments and Ablations:} We conduct thorough evaluations comparing variants of our system and baselines. Our results demonstrate the efficacy of each component and highlight the benefits of integrated explanations, as evidenced by improved success rates and highly faithful rationales.
\end{enumerate}

Our work takes a step toward navigation systems that are both adaptable to new domains and transparent in their operation. We hope this approach and the accompanying benchmark will spur further research in computer vision, natural language processing, and robotics for explainable embodied AI.

\section{Related Works}
\textbf{Sensor Fusion and Classical Navigation:} Combining multiple sensors to improve state estimation and navigation has a long history in robotics. The Kalman filter \cite{kalman1960} and its variants are core tools for fusing data from cameras, lidar, radar, and IMUs to estimate an agent’s pose and detect obstacles. Classical navigation systems (e.g., autonomous driving stacks or SLAM-based robot planners) use such estimates to plan safe trajectories. These systems excel in reliability and have well-understood failure modes, but they typically rely on hand-crafted rules or simplistic classifiers for scene understanding. For instance, a rule-based autonomous vehicle might react to a pedestrian crossing based on lidar blob detection and pre-defined distance thresholds, without semantic interpretation of the situation. Our work retains the strengths of sensor fusion for accurate geometry and tracking, while addressing its limitations by introducing a learned vision-language module to interpret and reason about complex scenes. One example is understanding that a person with an outstretched arm might be signaling to stop, something a pure geometric logic would miss.

\medskip

\textbf{Vision-Language Models for Embodied Agents:} Large-scale vision-language models have recently shown remarkable generalization in image understanding and reasoning. CLIP \cite{radford2021clip} demonstrated that jointly learning from images and their text descriptions yields powerful visual representations that transfer to many tasks. PaLM-E \cite{driess2023palme} extended this idea to embodiment, by feeding visual and continuous state information into a pre-trained language model, enabling it to perform tasks like planning and question answering directly from raw sensor inputs. Similarly, OpenAI’s GPT-4V \cite{openai2023gpt4} can accept images and text, exhibiting impressive zero-shot reasoning about visual scenes. In robotics, these models have been explored to some extent \cite{shah2022lmnav}. For example, \emph{SayCan} (Ahn \emph{et al.}, 2022) used language models to plan high-level actions from instructions, and more directly, Zawalski \emph{et al.} \cite{zawalski2024ecot} introduced Embodied Chain-of-Thought reasoning in a vision-language-action (VLA) model. Their approach trained a policy to produce intermediate textual reasoning steps (e.g., identifying objects, considering sub-goals) before outputting an action, which significantly improved generalization. We build on these insights by using a VLM backbone to handle perceptual understanding and reasoning in our navigation system. Unlike previous works that usually focus on a single domain or modality (e.g., vision-only input), our model ingests a rich set of sensors (images, depth, radar, etc.) by converting them into a unified vision-language representation, a derivation of PaLM-E's 'multimodal sentences' \cite{driess2023palme}. This allows the pretrained language model to ground its words and thinking in real-world sensor data.

\medskip

\textbf{Explainability and Chain-of-Thought in Navigation:} Providing explanations for robot actions has been studied in various forms. Kim \emph{et al.} \cite{kim2018bddx} introduced the idea of introspective explanations for self-driving cars, using an attention-based controller whose attention maps could be aligned with an explanation generator. In their BDD-X dataset and model, the network would output text such as ``The car slows down because a pedestrian is crossing'' while also highlighting the pedestrian in the camera view, thus giving users a rationale grounded in the model’s perception \cite{kim2018bddx}. Our approach shares this goal of aligning explanations with the true decision factors, but we leverage a much more expressive language model to generate \emph{chain-of-thought} style explanations. Chain-of-thought prompting \cite{wei2022cot} has proven effective for letting models articulate multi-step reasoning (e.g., in question answering tasks). In robotics, beyond the aforementioned ECoT approach \cite{zawalski2024ecot}, there is growing interest in using natural language as a medium for a robot to explain or even plan its actions in dialogue with humans. The SNEI dataset introduced by Liu \emph{et al.} \cite{liu2025snei} is a recent effort in this direction: it provides human-annotated visual question-answer pairs and chain-of-thought narratives for social navigation scenarios. They fine-tuned a vision-language assistant (Social-LLaVA) on SNEI, and showed it could produce high-level navigation instructions and explanations that were preferred by humans over those from GPT-4V in certain social scenarios. Unlike \cite{liu2025snei}, which focuses on interactive question answering and high-level advice for social robots, our work integrates explanation generation directly into the control loop of the agent and evaluates it on actual navigation performance. Moreover, we tackle multiple domains in one unified model, whereas prior explainable navigation works have been domain-specific (driving-only or social-only). We also introduce an explanation fidelity metric and training loss. Rather than just aligning attention \cite{kim2018bddx}, we explicitly check if the content of the explanation truly supports the chosen action by perturbing inputs. This is inspired by fidelty metrics in explainable AI literature \cite{deyoung2020eraser, jacovi2020faithfulness}, such as input erasure tests.

\medskip

\textbf{Benchmarks for Navigation and XAI:} Embodied AI has a variety of benchmarks, each targeting different aspects. For instruction-following indoor navigation, Room-to-Room (R2R) \cite{anderson2018r2r} is a seminal benchmark that uses the Matterport3D environments \cite{chang2017matterport3d} to evaluate an agent’s ability to follow natural language directions. R2R and its successors (RxR, R4R, etc.) focus on language understanding and vision, but do not require the agent to explain its decisions. In autonomous driving, large-scale datasets like nuScenes \cite{caesar2020nuscenes} and the Waymo Open Dataset \cite{sun2020waymo} provide multimodal sensor data (camera, lidar, radar) with annotations for perception tasks and even trajectory planning, but they lack textual explanations of the vehicle’s actions. The BDD-X dataset \cite{kim2018bddx} filled this gap by collecting human explanations for short driving video clips, and the Talk2Car dataset (Kim \emph{et al.}, 2020) introduced language commands to autonomous vehicles, but again no single benchmark evaluates both task performance and explanation quality. Existing benchmarks are also specialized by context. Our proposed benchmark is unique in its breadth and focus: it spans multiple domains (indoor, outdoor, social) and includes ground-truth annotations for both navigation and explainability. It enables, for the first time, a unified evaluation of an agent’s ability to adapt (perform well in different environments) and to explain (justify its actions in language).

\section{PhysNav-DG Architecture Methodology}

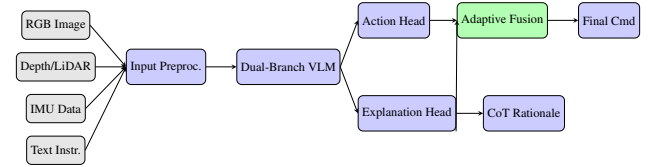
\begin{figure}[htbp]
    \centering
    \resizebox{\linewidth}{!}{
    \begin{tikzpicture}[node distance=0.365cm]
        \tikzset{
            block/.style = {rectangle, rounded corners, minimum width=1cm, minimum height=1cm, text centered, draw=black, fill=blue!20},
            arrow/.style = {thick,->,>=stealth},
            sensor/.style = {rectangle, rounded corners, minimum width=0.8cm, minimum height=0.8cm, text centered, draw=black, fill=gray!20},
            fusion/.style = {rectangle, rounded corners, minimum width=1cm, minimum height=1cm, text centered, draw=black, fill=green!30}
        }
        \node (rgb) [sensor] {RGB Image};
        \node (depth) [sensor, below=of rgb] {Depth/LiDAR};
        \node (imu) [sensor, below=of depth] {IMU Data};
        \node (text) [sensor, below=of imu] {Text Instr.};

        \node (preproc) [block, right=of depth, xshift=0.5cm] {Input Preproc.};

        \draw [arrow] (rgb.east) -- (preproc.west);
        \draw [arrow] (depth.east) -- (preproc.west);
        \draw [arrow] (imu.east) -- (preproc.west);
        \draw [arrow] (text.east) -- (preproc.west);

        \node (vlm) [block, right=of preproc, xshift=0.5cm] {Dual-Branch VLM};
        \draw [arrow] (preproc.east) -- (vlm.west);

        \node (action) [block, above right=0.3cm and 0.5cm of vlm] {Action Head};
        \node (explain) [block, below right=0.3cm and 0.5cm of vlm] {Explanation Head};

        \node (rationale) [block, right=of explain, xshift=0.4cm] {CoT Rationale};
        \node (fusion) [fusion, right=of action, xshift=0.4cm] {Adaptive Fusion};

        \node (final) [block, right=of fusion, xshift=0.5cm] {Final Cmd};

        \draw [arrow] (vlm.east) -- (action.west);
        \draw [arrow] (vlm.east) -- (explain.west);
        \draw [arrow] (action.east) -- (fusion.west);
        \draw [arrow] (explain.east) -- (rationale.west);
        \draw [arrow] (explain.south east) -- (fusion.west);
        \draw [arrow] (fusion.east) -- (final.west);
    \end{tikzpicture}
    }
    \caption{System Overview: Multiple sensor inputs are processed via Input Preprocessing and fed into a Dual-Branch VLM. The Action Head and Explanation Head work in parallel, with the Explanation Head contributing to Adaptive Fusion for final command generation.}
    \label{fig:old_overview}
\end{figure}

Figure \ref{fig:old_overview} provides an overview of the PhysNav-DG framework. Multiple sensor inputs are first processed via an input preprocessing module, then fed into a Dual-Branch Vision-Language Model (VLM) that predicts both navigation actions and chain-of-thought explanations. The generated explanations contribute to an adaptive fusion mechanism that combines classical sensor fusion with the semantic reasoning of the VLM.

\subsection{Input Preprocessing \& Multi-Sensor Feature Extraction}
To ensure robust navigation across diverse environments, all sensor inputs are preprocessed into a unified, tokenized format. Our preprocessing pipeline includes two complementary strategies for non-RGB sensor data:
\begin{itemize}
    \item \textbf{Visual Data:} Raw RGB images are processed using a pre-trained vision transformer (employing CLIP’s visual encoder) to produce a sequence of feature tokens that preserve spatial context. Depth images from stereo cameras and LiDAR point clouds are first converted into 2D depth maps. In parallel, raw sensor readings, such as accelerometer data, are rendered into visual plots (e.g., waveforms or spectrograms), following the methodology of Yoon et al. in \textit{By My Eyes}. This dual approach leverages both standard depth encoding and a compact visual representation of dynamic sensor data, reducing prompt length while enhancing the model’s capacity to capture complex sensor patterns.
    
    \item \textbf{Non-Visual Sensor Data:} Numeric measurements from IMUs, GPS, accelerometers, and other sources are normalized and discretized into concise text tokens. For example, the current speed and heading are encoded as 
    \verb|Speed=1.2 m/s, Heading=90° (east)|, 
    ensuring that all quantitative data are directly interpretable by the language model.
    
    \item \textbf{Instruction Encoding:} External navigation instructions (e.g., “Go down the hallway and turn left at the kitchen”) are appended as text, forming part of a multimodal prompt.
\end{itemize}

The final unified prompt combines visual tokens, textual sensor descriptors, and instructions, enabling our VLM to process heterogeneous data within a single input sequence.

\subsection{Dual-Branch Vision-Language Module}
Central to PhysNav-DG is a dual-branch VLM built on a transformer-based backbone (initialized from models such as PaLM‑E or GPT‑4V). The model is fine-tuned on an extensive corpus of navigation data to jointly predict precise navigation actions and generate chain-of-thought explanations in natural language.
\begin{itemize}
    \item \textbf{Action Prediction Branch:} The shared latent representation is directed to a dedicated head that outputs the next navigation command. For indoor navigation, this branch operates as a classifier over discrete actions (e.g., \texttt{MOVE\_FORWARD}, \texttt{TURN\_LEFT}, \texttt{TURN\_RIGHT}), while in driving scenarios it produces continuous control parameters (e.g., steering angle, throttle). Training is conducted using cross-entropy loss for discrete tasks and mean-squared error loss for continuous tasks, with ground-truth actions provided by human experts or high-fidelity simulation.
    
    \item \textbf{Explanation Generation Branch:} In parallel, the model generates a natural language chain-of-thought explanation detailing the reasoning behind the predicted action. For instance, the model might produce: 
    \begin{quote}
        “The sensor data indicates that the path ahead is clear, but an obstacle is detected on the left; thus, turning right minimizes collision risk.”
    \end{quote}
    Supervised training is employed using human-annotated or synthetically generated explanations. To enforce semantic alignment between the predicted action and the generated explanation, we introduce a consistency loss during training. After generating the chain-of-thought explanation, the model re-encodes it along with the original sensory context and compares the re-inferred action against the initially predicted action. We compute the consistency loss as a cross-entropy or mean-squared error (depending on the action space) between the original and explanation-conditioned predictions. This encourages explanations that genuinely reflect the model's internal rationale rather than superficial justifications. Additionally, we weight this loss term dynamically based on the confidence score of the explanation, ensuring it contributes proportionally to the training signal.
    
    \item \textbf{Two-Stage Decoding:} To enhance output quality, a two-stage decoding process is adopted \cite{yao2023react}. Initially, the model generates an internal chain-of-thought (prompted by a cue such as “Think step by step:”); the final action is then conditioned on this internal reasoning. Few-shot exemplars are provided during training to prime the model for producing outputs in the desired format.
\end{itemize}

\subsection{Adaptive Sensor Fusion and Confidence Estimation}

To combine the semantic reasoning capabilities of the VLM with the robust state estimation of classical sensor fusion, we incorporate a modified Adaptive Kalman Filter (AKF). A critical aspect of this fusion mechanism is the estimation of confidence from both the VLM and the AKF, which determines fusion dominance.

Let \(a_{\text{VLM}}\) and \(p_{\text{VLM}}\) denote the action and confidence predicted by the VLM, and let \(a_{\text{AKF}}\) and \(p_{\text{AKF}}\) represent the corresponding estimates from the AKF module. A dynamic weighting factor \(\alpha_t\) is computed as:
\[
\alpha_t = \frac{p_{\text{VLM}}}{p_{\text{VLM}} + p_{\text{AKF}}}.
\]

\subsubsection{Confidence Estimation in the VLM}
When employing advanced models such as LLaMA 3.2, confidence estimation is multifaceted:
\begin{itemize}
    \item \textbf{Action Probability Distributions:} For discrete tasks, the model produces a softmax probability distribution over actions (e.g., \texttt{MOVE\_FORWARD}, \texttt{TURN\_LEFT}, \texttt{TURN\_RIGHT}). The probability of the selected action directly serves as \( p_{\text{VLM}} \). In continuous control settings, the regression outputs include uncertainty measures (such as predictive variance) \cite{kendall2017uncertainties}
that are mapped to a confidence score.
    
    \item \textbf{Chain-of-Thought Consistency:} The internal chain-of-thought (CoT) explanation generated by the model is analyzed for consistency with the final action. A coherent and logically sound CoT reinforces the confidence score.
    
    \item \textbf{Self-Evaluation and Calibration:} LLaMA 3.2 11B can be fine-tuned to provide self-assessment regarding its predictions. If the model identifies ambiguous or low-quality inputs (e.g., cluttered visuals or low contrast), this uncertainty is incorporated into \( p_{\text{VLM}} \) using calibration methods such as temperature scaling. \cite{guo2017calibration}
    
    \item \textbf{Erasure-Based Verification:} As an external validation, key visual features referenced in the explanation are selectively masked (e.g., by blurring or occlusion) \cite{fong2017perturb}. A significant drop in confidence upon erasure confirms the importance of those features and validates the explanation. Conversely, if the drop is negligible, \( p_{\text{VLM}} \) is reduced to mitigate potential overconfidence.
\end{itemize}

\subsubsection{Confidence Estimation in the AKF}
The classical sensor fusion module (AKF) provides a complementary confidence measure:
\begin{itemize}
    \item \textbf{Error Covariance Analysis:} The AKF computes an error covariance matrix reflecting the uncertainty in its state estimations. A lower covariance indicates higher confidence in the sensor data, yielding a larger \( p_{\text{AKF}} \).
    
    \item \textbf{Contextual Adaptation:} The AKF dynamically adjusts its process and measurement noise based on environmental conditions (e.g., lighting, weather) and sensor reliability (e.g., GPS integrity, IMU performance), ensuring that \( p_{\text{AKF}} \) accurately mirrors the current conditions.
\end{itemize}

\subsubsection{Fusion of Navigation Commands}
The final navigation decision is obtained by fusing the VLM and AKF outputs:
\begin{itemize}
    \item \textbf{For Continuous Control Commands:}
    \[
    a_{\text{fusion}} = \alpha_t \, a_{\text{VLM}} + (1-\alpha_t) \, a_{\text{AKF}},
    \]
    
    \item \textbf{For Discrete Action Spaces:}
    \[
    a_{\text{fusion}} =
    \begin{cases}
    a_{\text{VLM}}, & \text{if } \alpha_t > 0.5,\\[1mm]
    a_{\text{AKF}}, & \text{otherwise}.
    \end{cases}
    \]
\end{itemize}

Figure~\ref{fig:adaptive_sensor_fusion} visually illustrates two scenarios: (a) when the VLM exhibits high confidence (thus dominating the fusion), and (b) when the AKF’s confidence is high, leading to a fusion weighted towards the classical sensor fusion module.

\begin{figure}[htbp]
  \centering
  \begin{subfigure}[b]{0.45\textwidth}
    \centering
    \begin{tikzpicture}[scale=1, every node/.style={font=\small}]
      \draw[->] (0,0) -- (5,0) node[right]{Confidence};
      \draw[->] (0,0) -- (0,3) node[above]{Fusion Weight $\alpha_t$};
      
      \fill[blue!50] (1,0) rectangle (1.5,2.5);
      \node[rotate=90] at (1.25,1.25) {\textbf{VLM}};
      
      \fill[red!50] (3,0) rectangle (3.5,1);
      \node[rotate=90] at (3.25,0.5) {\textbf{AKF}};
      
      \draw[very thick,->,green!70!black] (0,2.13) -- (5,2.13)
        node[midway, above, yshift=4] {$\alpha_t \approx 0.71$};
      
      \node at (2.5, 3.7) {\textbf{(a) High VLM Confidence}};
    \end{tikzpicture}
    \caption{VLM dominates the fusion when its confidence is high.}
    \label{fig:adaptive_a}
  \end{subfigure}
  \hspace{0.05\textwidth}
  \begin{subfigure}[b]{0.45\textwidth}
    \centering
    \begin{tikzpicture}[scale=1, every node/.style={font=\small}]
      \draw[->] (0,0) -- (5,0) node[right]{Confidence};
      \draw[->] (0,0) -- (0,3) node[above]{Fusion Weight $\alpha_t$};
      
      \fill[blue!50] (1,0) rectangle (1.5,1);
      \node[rotate=90] at (1.25,0.5) {\textbf{VLM}};
      
      \fill[red!50] (3,0) rectangle (3.5,2.5);
      \node[rotate=90] at (3.25,1.25) {\textbf{AKF}};
      
      \draw[very thick,->,green!70!black] (0,0.87) -- (5,0.87)
        node[midway, above, yshift=4] {$\alpha_t \approx 0.29$};
      
      \node at (2.5, 3.7) {\textbf{(b) High AKF Confidence}};
    \end{tikzpicture}
    \caption{AKF dominates the fusion when its confidence is high.}
    \label{fig:adaptive_b}
  \end{subfigure}
  
  \vspace{0.5cm}
  \caption{Adaptive Sensor Fusion: Graphical depiction of how the fusion weight \(\alpha_t = \frac{p_{\text{VLM}}}{p_{\text{VLM}} + p_{\text{AKF}}}\) is dynamically adjusted based on the confidence levels from the VLM and AKF modules. In (a) high VLM confidence drives the decision, whereas in (b) high AKF confidence shifts the fusion toward the classical sensor fusion module.}
  \label{fig:adaptive_sensor_fusion}
\end{figure}
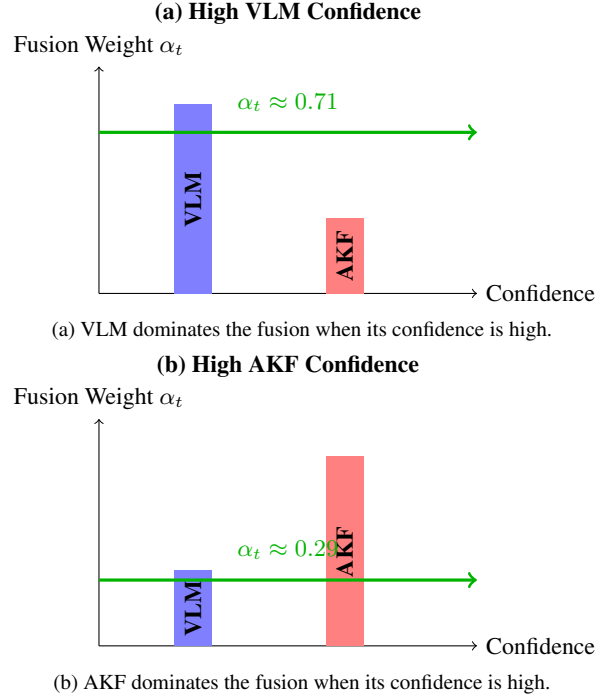

\begin{figure*}[h]
    \centering
    \begin{tabularx}{\textwidth}{|X|X|X|}
        \hline
        \textbf{Sensor Input} & \textbf{Action Taken} & \textbf{Explanation} \\ \hline
        GPS: Slight drift detected, Lidar: Obstacle 5m ahead & Reduce speed and adjust trajectory & "Slowing down due to an obstacle ahead and potential GPS drift. Verifying safe route." \\ \hline
        Camera: Pedestrian crossing detected & Full stop at crosswalk & "Halting as a pedestrian is detected in the crossing zone." \\ \hline
    \end{tabularx}
    \caption{Example of real-time explanations generated by PhysNav-DG alongside sensor inputs and actions.}
    \label{fig:cot_explanation}
\end{figure*}

In summary, the PhysNav-DG framework integrates classical sensor fusion with a dual-branch vision-language model, leveraging adaptive confidence estimation to achieve robust, domain-adaptive navigation with transparent, human-interpretable decision-making.

\section{MD-NEX Benchmark Methodology}
A major contribution of this work is the Multi-Domain Navigation and Explanation Benchmark (MD-NEX) we created. This section details how we curated the data, synthesized additional annotations, and defined the evaluation protocol.

\subsection{Domains and Data Sources}
We targeted three distinct domains to cover a broad spectrum:

\begin{enumerate}
    \item \textbf{Indoor Navigation:} We leverage the Matterport3D environment [9], which provides 3D scans of real buildings. We specifically build on the Room-to-Room (R2R) dataset [8] for navigation instructions in these environments. R2R offers 21,000+ crowd-sourced instructions for paths in 90 buildings. We utilize the same environments, but extend the tasks. Instead of only following instructions, we also consider autonomous exploration goals (e.g., “find the elevator”) and we require explanations. As R2R does not contain explanations, we generated them. Using GPT-4o-mini, we automatically created explanations for why each action in an instruction-following trajectory is taken. For example, if the instruction says “turn left at the end of the hall”, and the agent does so, we may generate the explanation “I turned left at the end of the hall as instructed, since the hall ended and the directions said to turn.” We also generated and integrated environment-driven explanations. If an agent has to deviate from instructions due to an obstacle, it may explain that “the path was blocked, so I had to go around”. The indoor data totals 9,782 episodes: 6,891 from R2R (with synthetic explanations) and 2,891 additional ones with other goals (some using Matterport’s building map to set random goals, with instructions synthesized by GPT-4o-mini to resemble human instructions). All episodes come with sensor streams: panoramic RGB-D images, compass, and ground-truth poses for evaluation.

    \item \textbf{Outdoor Driving:} We combine multiple datasets to capture driving. The base is BDD-X [7], which contains over 7,000 video clips of driving with human-written explanations for the driver’s actions. We augment BDD-X with sensor data from the nuScenes dataset [10] and Waymo Open Dataset [11]. Since nuScenes and Waymo provide lidar, radar, and multiple cameras for driving scenarios (but no explanations), we align a subset of BDD-X clips with similar scenarios in nuScenes (e.g., intersections, turns, etc.) and transfer the explanations. We employ a BLIP-2 image captioner and GPT-4o-mini driving commentary model to generate explanations for nuScenes/Waymo sequences. For instance, for a given nuScenes scene where the ego vehicle makes a right turn at a crosswalk, we may generate text: “The car turns right, slowing down to yield to a pedestrian crossing.” These are verified with human reviewers for correctness. The outdoor dataset contains 12,346 episodes: 7,023 from BDD-X with native explanations and 5,323 aligned sequences from nuScenes and Waymo datasets, augmented with BLIP-2 and GPT-4o-mini-generated rationales and verified by human reviewers. Each has a front camera video, 6 surrounding camera views (for 360°), lidar point clouds, radar scans (where available from nuScenes), and CAN bus info (speed, steering). We downsample longer sequences into shorter segments with one key decision requiring explanation per segment (e.g., a turn, lane change, or obstacle avoidance).

    \item \textbf{Social Navigation:} For navigating in crowds and social settings, we base our data on SNEI [12] and the Socially Compliant Navigation Dataset (SCAND, 2021). SNEI provides 40,000 visual question-answer pairs and chain-of-thought annotations about social scenarios, but not all are navigation episodes. We extract scenarios from SNEI where a robot is moving in a crowd (descriptions were turned into episodes). We also employ SCAND, containing real robot trajectories in crowds (no explanations). These trajectories are annotated by humans with explanations at noteworthy points (e.g., “Robot paused to let person pass, then continued behind them.”). We employed the SocialForce model to simulate pedestrians to generate a further 2,000 episodes, for which explanations were generated using GPT-4o-mini and verified by human annotators.The social navigation set contains a total of 5,218 episodes. Each episode has multiple humans moving with some social context (like a group talking in a hallway, or a person carrying something fragile that the robot should be extra careful around – these contexts are described to the agent in a textual form as well (e.g., “Context: person with fragile object ahead”).
\end{enumerate}

\subsection{Data Aggregation and Quality Control}
We found that exclusively using existing datasets was insufficient to cover all desired domains with explanations. Thus, we web-scraped additional data from sites like Flickr and Unsplash for diverse scenes. This includes images of indoor clutter, unusual driving scenes (construction zones, emergency vehicles), and dense crowds. From these images, we either created single-step scenarios or integrated them into sequences. We automatically captioned these images with BLIP-2 to get a basic description, then employed a GPT-4o-mini model to assign a plausible action and explanation (e.g., Figure 4). 

For example, a photo of a dog crossing a street might be turned into a scenario: state: front camera image with dog, action: brake, explanation: “Stopping because a dog is crossing the road.” These single-step snippets (of which there are 3,500) were used to augment training, improving the model’s knowledge of rare situations. Augmentation (reference figure) was applied to all images to 5x sample count, increasing to 17,500.

\begin{figure}[h]
    \centering
    \begin{tikzpicture}[
        font=\footnotesize,
        block/.style={
            rectangle,
            draw,
            rounded corners,
            align=left,
            text width=1.5cm,   
            minimum height=1cm
        }
    ]
        \node[anchor=south west, inner sep=0] (img) at (0,0)
        {\includegraphics[width=0.25\textwidth,height=3.5cm,keepaspectratio=false]{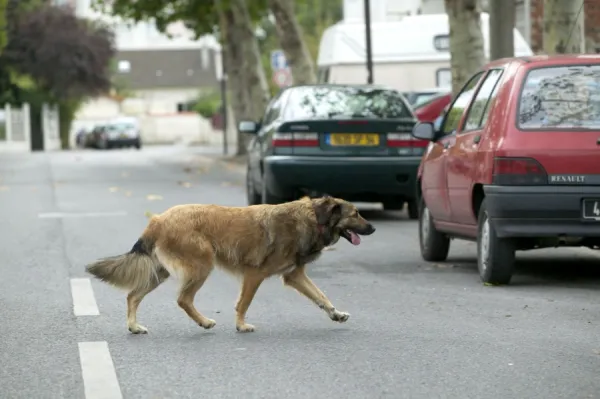}};
        
        \node[block,
              anchor=north west,
              xshift=1.2cm,
              yshift=-0.3cm
             ] (action) at (img.north east)
        {%
            \textbf{Action:} Brake
        };
        
        \node[
            rectangle,
            draw,
            rounded corners,
            align=left,
            text width=2.5cm,      
            minimum height=1cm,
            below=0.5cm of action
        ] (explanation)
        {%
            \textbf{Explanation:}\\
            "Stopping because a dog is crossing the road."
        };
        
        \draw[->, thick, shorten >=1pt] (img.east |- action.west) -- (action.west);
        \draw[->, thick, shorten >=1pt] (action.south) -- (explanation.north);
    \end{tikzpicture}
    
    \caption{An example single-step scenario. The front camera detects a dog crossing the street, prompting the system to brake. The explanation is provided to clarify the rationale for the action.}
    \label{fig:dog_scenario}
\end{figure}
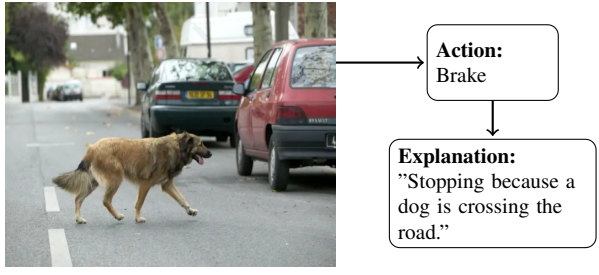

All generated explanations, whether via language model or script, were reviewed by human annotators for correctness and clarity. If an explanation was incorrect (e.g., “car stops for red light” but the light was actually green), we fixed or discarded it. 

In total, MD-NEX contains 30,936 unaugmented navigation episodes, each with multi-modal sensor data, ground-truth actions, and human-verified or synthetic explanations. 

\subsection{Evaluation Metrics}
We evaluate navigation performance and explanation quality separately, and also consider a combined score.

\begin{enumerate}
    \item \textbf{Success Rate (SR)}: the fraction of episodes where the agent successfully reaches the goal or completes the required task. 
    Criteria for success vary: in indoor, reaching the target location within a tolerance; in driving, completing the route without infractions; in social, reaching destination without collisions or rule violations. 

    \item \textbf{Success weighted by Path Length (SPL)}: a standard VLN metric [8] that penalizes inefficiency. It is defined as $S \frac{L_{opt}}{L_{agent}}$ where $S$ is a binary success indicator, $L_{opt}$ is the shortest path length to goal, and $L_{agent}$ is the path length taken by agent. We report average SPL across episodes, which reflects both success and efficiency.

    \item \textbf{Collision Rate}: number of collisions per meter or per episode. We compute this in simulation by checking overlaps; for driving, we also treat major infractions (like running into a pedestrian) as collisions. We also measure specifically pedestrian collisions in social domain as a subset.

    \item \textbf{Explanation Faithfulness}: how well the explanation reflects the true causes of the decision. We employ an erasure-based metric inspired by prior work in explainable NLP \cite{deyoung2020eraser, jacovi2020faithfulness}. For each decision, we remove or mask the top-$k$ features that the explanation highlights (these could be image regions or tokens) and see if the VLM’s action prediction changes. We then compute a score per explanation:
: 

    \[
    F = 1 - \frac{\text{performance when feature removed}}{\text{performance normally}}
    \]
    
    This represents the fractional drop in confidence or correctness. If removing the explained features causes the model to make a different decision (or lowers its confidence significantly), the explanation is deemed faithful (score near 1). If removal has no effect, score near 0. We average these over the test set.

    \item \textbf{Readability}: We assess explanation clarity and grammar. Using a combination of automated metrics and human ratings, we measure this on a 5-point Likert scale (5 = perfectly clear and fluent, 1 = incomprehensible). We also verify that the reading level is appropriate: using a Flesch-Kincaid grade analysis.

    \item \textbf{Combined Score}: We report an overall score that multiples success rate with explanation faithfulness (to reward agents that succeed and explain well) for model comparison.
\end{enumerate}

\section{Results}

\begin{table*}[ht]
\centering
\caption{Performance of Navigation and Explanation on the MD-NEX Benchmark}
\label{tab:results}
\begin{tabular}{lccccccc}
\toprule
\textbf{Method} & \textbf{Domain} & \textbf{SR (\%)} & \textbf{SPL} & \textbf{Collision Rate} & \textbf{Exp. Faithfulness} & \textbf{Readability (5-pt)} & \textbf{Combined Score} \\
\midrule
\multirow{3}{*}{Classical} 
    & Indoor  & 73  & 0.62   & 0.10   & N/A   & N/A  & N/A \\
    & Outdoor & 68  & 0.60   & 0.08   & N/A   & N/A  & N/A \\
    & Social  & 65  & 0.58   & 0.09   & N/A   & N/A  & N/A \\
\midrule
\multirow{3}{*}{Single-Branch} 
    & Indoor  & 65  & 0.56   & 0.06   & 0.75 & 4.0 & 0.49 \\
    & Outdoor & 80  & 0.66   & 0.04   & 0.78 & 4.2 & 0.62 \\
    & Social  & 70  & 0.54   & 0.05   & 0.74 & 4.0 & 0.52 \\
\midrule
\multirow{3}{*}{No Fusion} 
    & Indoor  & 67  & 0.59   & 0.05   & 0.83 & 4.4 & 0.56 \\
    & Outdoor & 82  & 0.68   & 0.03   & 0.82 & 4.5 & 0.67 \\
    & Social  & 73  & 0.57   & 0.04   & 0.80 & 4.3 & 0.58 \\
\midrule
\multirow{3}{*}{Proposed} 
    & Indoor  & 78  & 0.67   & 0.04   & 0.87 & 4.5 & 0.68 \\
    & Outdoor & 91  & 0.75   & 0.02   & 0.87 & 4.6 & 0.79 \\
    & Social  & 83  & 0.65   & 0.03   & 0.86 & 4.5 & 0.71 \\
\bottomrule
\end{tabular}
\end{table*}

We evaluate our integrated system (Dual-Branch VLM + Adaptive Kalman Filter) against several baselines and ablations. Table~\ref{tab:results} summarizes the performance of four methods evaluated on the MD-NEX Benchmark across three domains: Indoor, Outdoor, and Social Navigation. The compared methods are:

\begin{enumerate}
    \item \textbf{Classical:} A traditional sensor fusion approach using an Adaptive Kalman Filter (AKF) combined with rule-based planning, without any learning-based vision-language integration or explanation generation.
    \item \textbf{Single-Branch:} An end-to-end vision-language model (VLM) architecture that produces a unified output combining navigation action and explanation in a single decoding process. This approach leverages high-level semantic reasoning but lacks explicit separation of action prediction and explanation.
    \item \textbf{No Fusion:} A dual-branch VLM that generates separate outputs for navigation actions and explanations. However, this variant does not incorporate the adaptive fusion mechanism with the classical sensor module, which limits its ability to correct for sensor noise or ambiguities in the visual input.
    \item \textbf{Proposed:} Our full model, integrating a dual-branch VLM with an adaptive sensor fusion mechanism. This model dynamically balances the VLM’s semantic outputs with classical state estimation, effectively bridging the gap between high-level semantic reasoning and geometric planning.
\end{enumerate}

\subsection{Error Analysis}
Our error analysis revealed several key failure modes in our proposed system. In challenging lighting conditions or during adverse weather, the visual encoder sometimes failed to extract reliable features, leading to incorrect or ambiguous actions and explanations. Additionally, a subset of cases had inconsistencies between the generated chain-of-thought and the final action, suggesting that the consistency loss occasionally did not fully align the two branches. We also observed that in highly dynamic environments, particularly during rapid pedestrian movements in social navigation scenarios, the adaptive fusion mechanism sometimes misjudged the confidence levels. This resulted in suboptimal blending of the VLM and classical sensor outputs. These failures indicate that while our system performs robustly overall, improvements in sensor data preprocessing, confidence calibration, and real-time adaptation are necessary to mitigate errors in edge cases.

\subsection{Discussion}
The experimental results clearly demonstrate that the integration of a dual-branch VLM with adaptive sensor fusion (Proposed) leads to substantial performance gains in both navigation accuracy and explanation quality. Specifically:
\begin{itemize}
    \item \textbf{Navigation Performance:} The proposed method achieves significantly higher success rates and efficiency (as measured by SPL) across all domains, reducing collisions markedly compared to the Classical baseline.
    \item \textbf{Explanation Quality:} The model generates explanations that are both more faithful and more readable than those produced by the Single-Branch and No Fusion variants. Decoupling action prediction from explanation generation, while reinforcing consistency between them, demonstrates the benefits of our dual-branch design.
    \item \textbf{Safety and Robustness:} The adaptive fusion mechanism effectively mitigates the limitations of purely learning-based approaches by dynamically incorporating classical sensor fusion outputs. This is critical in scenarios where visual ambiguity or environmental noise might lead to suboptimal decisions.
    \item \textbf{Cross-Domain Generalizability:} Our results demonstrate that the Proposed method consistently maintains high performance across diverse domains. With success rates of 78\%, 91\%, and 83\% in Indoor, Outdoor, and Social scenarios, respectively, and Combined Scores ranging from 0.68 to 0.79, the system shows robust generalizability. In contrast, the Classical approach exhibits inconsistent performance with marked drops across different domains.
\end{itemize}

The key takeaway is that our approach bridges the gap between high-level semantic reasoning and geometric planning. By fusing the strengths of learning-based VLMs with the precision of classical sensor fusion, our system not only improves navigation accuracy and safety but also enhances the transparency of decision-making through real-time chain-of-thought explanations. This integrated framework paves the way for safer, more interpretable autonomous systems in applications such as autonomous driving, search-and-rescue robotics, and industrial automation.

\section{Conclusions and Future Works}
Our work introduces a promising bridge between high-level semantic reasoning and geometric planning \cite{chakraborti2019xaip, doshivelez2017definitions} by integrating a novel dual-branch vision-language model with adaptive sensor fusion. This hybrid approach leverages the interpretability and rich contextual understanding of state-of-the-art VLMs alongside the precision and robustness of classical sensor fusion, yielding significant improvements in navigation accuracy, efficiency, and safety across diverse domains. By generating coherent chain-of-thought explanations, our method enhances transparency, thereby fostering greater trust in autonomous systems operating in complex and dynamic environments.

Future research should focus on refining confidence estimation and calibration techniques to improve the adaptive fusion process, particularly in scenarios characterized by ambiguous sensor data or rapid environmental changes\cite{vishen2025advancing, hooker2019benchmark, adebayo2018sanity}. Expanding our multi-domain benchmark to cover additional real-world scenarios, such as industry automation and search-and-rescue operations, will also be crucial in advancing the generalizability and reliability of explainable navigation systems.
{
    \small
    \bibliographystyle{ieeenat_fullname}
    \bibliography{sec/bibliography}
}


\end{document}